\documentclass[conference]{IEEEtran}
\IEEEoverridecommandlockouts
\usepackage[acronym]{glossaries}
\usepackage[hidelinks]{hyperref}

\usepackage{cite}
\usepackage{amsmath,amssymb,amsfonts}
\usepackage{algorithmic}
\usepackage{textcomp}
\usepackage{xcolor}
\def\BibTeX{{\rm B\kern-.05em{\sc i\kern-.025em b}\kern-.08em
    T\kern-.1667em\lower.7ex\hbox{E}\kern-.125emX}}

\usepackage{graphicx}
\usepackage{booktabs}
\usepackage{adjustbox}
\usepackage{svg}
\usepackage{soul}

\IEEEoverridecommandlockouts

\makeglossaries
\newacronym{ai}{AI}{Artificial Intelligence}
\newacronym{aoi}{AOI}{Area of Interest}
\newacronym{bilstm}{BiLSTM}{Bidirectional Long Short-Term Memory}
\newacronym{cot}{CoT}{Chain of Thought}
\newacronym{ct}{CT}{Conversation Tree}
\newacronym{cnn}{CNN}{Convolutional Neural Network}
\newacronym{cv}{CV}{Computer Vision}

\newacronym{drl}{DRL}{Deep Reinforcement Learning}
\newacronym{ddpg}{DDPG}{Deep Deterministic Policy Gradient}
\newacronym{dl}{DL}{Deep Learning}
\newacronym{dnn}{DNN}{Deep Neural Networks}
\newacronym{dqn}{DQN}{Deep Q-learning}
\newacronym{dpo}{DPO}{Direct Preference Optimization}
\newacronym{ddqn}{DDQN}{Double Q-learning}
\newacronym{et}{ET}{Eye-tracking}
\newacronym{ft}{FT}{Fine-Tuning}
\newacronym{gae}{GAE}{Generalized Advantage Estimate}
\newacronym{gnn}{GNN}{Graph Neural Networks}
\newacronym{gqa}{GQA}{Grouped Query Attention}
\newacronym{hrlaif}{HRLAIF}{Hybrid Reinforcement Learning from AI Feedback}
\newacronym{il}{IL}{Imitation Learning}
\newacronym{ipopt}{IPOPT}{Interior Point Optimizer}
\newacronym{kl}{KL}{Kullback–Leibler}
\newacronym{lm}{LM}{Language Model}
\newacronym{lms}{LMs}{Language Models}
\newacronym{llm}{LLM}{Large Language Model}
\newacronym{llms}{LLMs}{Large Language Models}
\newacronym{lstm}{LSTM}{Long short-term memory}
\newacronym{lora}{LoRA}{Low-Rank Adaptation}
\newacronym{lr}{LR}{Learning Rate}
\newacronym{mdp}{MDP}{Markov Decision Process}
\newacronym{mha}{MHA}{multi-head attention}
\newacronym{ml}{ML}{Machine Learning}
\newacronym{mllm}{MLLM}{Multimodal Large Language Model}
\newacronym{mllms}{MLLMs}{Multimodal Large Language Models}
\newacronym{mlp}{MLP}{Multilayer Perceptron}
\newacronym{mpnn}{MPNN}{Message Passing Neural Networks}
\newacronym{mtl}{MTL}{Multi-task Learning}
\newacronym{ner}{NER}{Named Entity Recognition}
\newacronym{nlp}{NLP}{Natural Language Processing}
\newacronym{nn}{NN}{Neural Networks}
\newacronym{ocr}{OCR}{Optical Character Recognition}
\newacronym{orms}{ORMs}{Outcome-supervised reward models}
\newacronym{p3o}{P3O}{Pairwise Proximal Policy Optimization}
\newacronym{peft}{PEFT}{Parameter-Efficient Fine-Tuning}
\newacronym{pi}{PI}{Policy Iteration}
\newacronym{ppo}{PPO}{Proximal Policy Optimization}
\newacronym{prms}{PRMs}{Process Based Reward Models}
\newacronym{qa}{QA}{Question Answering}
\newacronym{raft}{RAFT}{Reward rAnked FineTuning}
\newacronym{rag}{RAG}{Retrieval-Augmented Generation}
\newacronym{rnn}{RNN}{Recurrent Neural Network}
\newacronym{rnns}{RNNs}{Recurrent Neural Networks}
\newacronym{rl}{RL}{Reinforcement Learning}
\newacronym{rlaif}{RLAIF}{Reinforcement Learning from AI Feedback}
\newacronym{rlcd}{RLCD}{Reinforcement Learning from Contrastive Distillation}
\newacronym{rlhf}{RLHF}{Reinforcement Learning from Human Feedback}
\newacronym{rm}{RM}{Reward Model}
\newacronym{rms}{RMs}{Reward Models}
\newacronym{rso}{RSO}{Statistical Rejection Sampling Optimization}
\newacronym{rrhf}{RRHF}{Rank Responses to align Human Feedback}

\newacronym{sota}{SOTA}{State of the Art}
\newacronym{sft}{SFT}{Supervised Fine-Tuning}
\newacronym{smoe}{SMoE}{Sparse Mixture of Experts}
\newacronym{slic}{SLiC}{Sequence Likelihood Calibration}
\newacronym{slcihf}{SLiC-HF}{Sequence Likelihood Calibration with Human Feedback}
\newacronym{sorms}{SORMs}{Stepwise ORMs}
\newacronym{swa}{SWA}{sliding window attention}
\newacronym{trpo}{TRPO}{Trust Region Policy Optimization}
\newacronym{trt}{TRT}{total reading time}

\newacronym{vi}{VI}{Value Iteration}
\newacronym{vqa}{VQA}{Visual Question Answering}

\begin{document}

\title{Integrating Cognitive Processing Signals into Language Models: A Review of Advances, Applications and Future Directions. \thanks{© 2025 IEEE. This is the author's accepted manuscript. The final version will appear in IEEE under the Proceedings of the International Joint Conference on Neural Networks 2025.}}


\author{
    \IEEEauthorblockN{Angela Lopez-Cardona\IEEEauthorrefmark{1} \IEEEauthorrefmark{2}, Sebastian Idesis\IEEEauthorrefmark{1}, Ioannis Arapakis\IEEEauthorrefmark{1}}
    \IEEEauthorblockA{\IEEEauthorrefmark{1}Telef\'{o}nica Scientific Research, Barcelona, Spain\\
    \IEEEauthorrefmark{2}Universitat Politècnica de Catalunya, Barcelona, Spain\\
    Email: \{angela.lopez.cardona, sebastianariel.idesis, ioannis.arapakis\}@telefonica.com}
}

\maketitle

\begin{abstract}
Recently, the integration of cognitive neuroscience in \acrfull{nlp} has gained significant attention. This article provides a critical and timely overview of recent advancements in leveraging cognitive signals, particularly \acrfull{et} signals, to enhance \acrfull{lms} and \acrfull{mllms}. By incorporating user-centric cognitive signals, these approaches address key challenges, including data scarcity and the environmental costs of training large-scale models. Cognitive signals enable efficient data augmentation, faster convergence, and improved human alignment. The review emphasises the potential of \acrshort{et} data in tasks like \acrfull{vqa} and mitigating hallucinations in \acrshort{mllms}, and concludes by discussing emerging challenges and research trends.
\end{abstract}

\begin{IEEEkeywords}
eye-tracking, cognitive signals, natural language processing, large language models, transformers
\end{IEEEkeywords}

\section{Introduction}
\label{sec:intro}

\acrfull{lms} have emerged as powerful tools for processing textual representations and modeling complex relationships between input features and output labels. Their applications span a wide range of tasks, including text generation, comprehension, translation, and sentiment analysis \cite{radford_improving_2018, devlin_bert_2018}. These developments in \acrfull{nlp} have been significantly accelerated by Transformer architectures \cite{vaswani_attention_2017}, which enabled the parallelization of input sequence processing via self-attention mechanisms that capture long-range dependencies effectively. 

Building on the foundation of Transformer architectures, models like BERT \cite{devlin_bert_2018} were proposed, leveraging Transformer encoders to capture bidirectional context within text and popularising the pretraining and fine-tuning paradigm. Similarly, Transformer decoders led to the development of models like GPT \cite{radford_improving_2018}, which framed language modeling as an autoregressive task
. These advancements contributed to the rapid rise of \acrfull{llms}, marking a milestone in \acrshort{nlp}. Popular examples such as GPT-4 \cite{openai_gpt-4_2023} and LLaMA 3 \cite{dubey_llama_2024} have become benchmarks in \acrfull{ai}, establishing the term ``generative language models'' due to their ability to produce coherent and human-like responses. 

More recently, \acrshort{llms} have evolved into \acrfull{mllms}, which extend their functionality beyond textual data, to train over heterogeneous datasets including images, audio, and video. This enables \acrshort{mllms} to process, understand, and generate content across various formats, further bridging the gap between data representations \cite{yin_survey_2024}. 

An emerging challenge in this context is the scarcity of sufficient data or the suboptimal use of existing data collections for training, which creates a significant bottleneck \cite{erdil_data_2024}. Current estimates suggest that the available stock of public human-generated text data is reaching saturation  \cite{villalobos_will_2024}. This calls for rethinking our training paradigm and researching effective synthetic data generation methods. Another important issue is the significant environmental cost associated with training large-scale models. As the size and complexity of \acrshort{llms} and \acrshort{mllms} continue to expand exponentially, their energy demands have surged, resulting in substantial CO$_{2}$ emissions. For example, while training the original Transformer model \cite{vaswani_attention_2017} resulted in an estimated emission of 0.012 tonnes of CO$_2$, the training of GPT-3 generated approximately 502 tonnes \cite{strubell2019, luccioni2022, Kaack2022}, a drastic increase across four orders of magnitude. 
These statistics highlight the pressing need for more sustainable and energy-efficient training practices. In addition to these considerations, aligning models with human values---such as the ability to follow instructions, avoid harmful behaviors, and engage in safe, user-friendly interactions---remains a significant challenge \cite{zhao_survey_2023, casper_open_2023}. Moreover, there is the persistent issue of models hallucinating, which leads to the generation of factually incorrect content \cite{yin_survey_2024, casper_open_2023}. 


A promising research direction is to enrich existing datasets with novel representations and user-derived cognitive signals. This approach can augment current data sources, reduce training data requirements, and enable faster convergence with more efficient learning for \acrshort{llms}. Moreover, such data augmentation methods hold the potential to improve human alignment and reduce hallucinations. One representative example is \acrfull{et}, which allows the study of visual attention and behaviour in a non-invasive and accurate manner. Reading is a complex process of information foraging that engages numerous cognitive functions \cite{bolliger_emtec_2024}, \cite{damelio_tpp-gaze_2024}. A substantial body of research has explored oculomotor behaviour in the context of reading comprehension and broader reading tasks, offering valuable insights into how attention patterns relate to human preferences and reward modeling \cite{mathias_survey_2020}. For example, \cite{adams_active_2015} demonstrated that oculomotor behavior reflects precision-weighted prediction error minimization within the framework of active inference. 
Similarly, models like UniAR \cite{li_uniar_2024} further highlight how visual attention patterns can predict explicit human feedback. 


The present article 
provides a focused and timely review of the most recent research that incorporates cognitive signals in language modelling and understanding, with a particular emphasis on the use of \acrshort{et} as a proxy for cognitive processes. Given the emergence of linguistic applications that require images, such as \acrfull{vqa}, and the integration of vision and language understanding in \acrshort{mllms}, we also consider the application of cognitive signals in image processing. However, a comprehensive review of the state of the art on \acrshort{llms} and \acrshort{mllms} extends beyond our scope, and for that we refer the readers to \cite{zhao_survey_2023} and \cite{yin_survey_2024}, respectively.

\begin{figure}[h]
    \begin{center}
    \includegraphics[width=\linewidth]{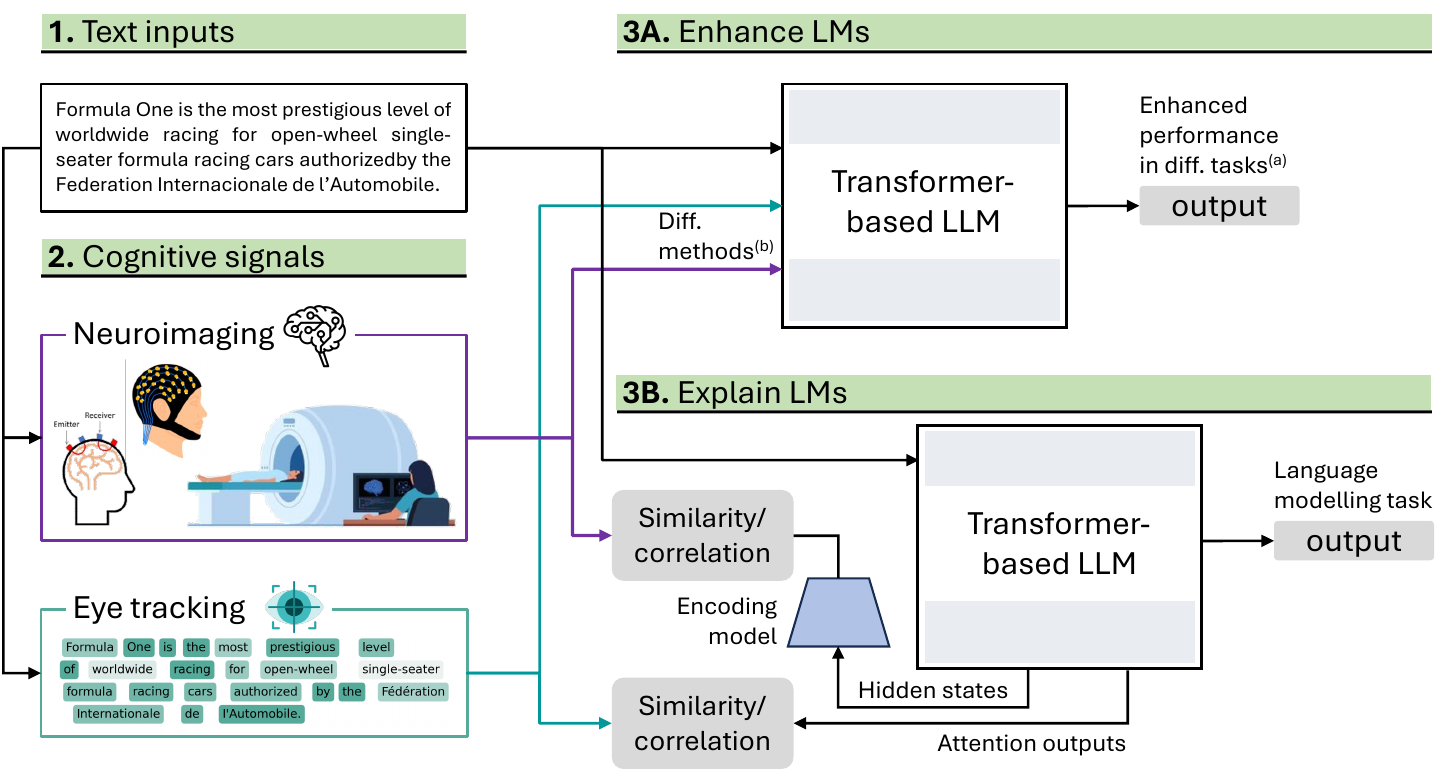}
    \end{center}
    \caption{Schematic overview illustrating the use of cognitive signals (2) to enhance \acrshort{lms} (3A) and to explore similar representations (3B). \textit{(a)}: applications are discussed in \autoref{sec:applications}, \textit{(b)}: integration methods in \autoref{sec:integration}.}
    \label{fig:summary}
\end{figure}

The structure of this article is organised as follows. First, in \autoref{sec:related_work}, we provide a comparative analysis of our work with existing reviews, highlighting key distinctions. In \autoref{sec:data}, we examine the modelling of various data modalities (\autoref{fig:summary}.2), followed by an exploration of their diverse applications in \autoref{sec:applications} (\autoref{fig:summary}.3A). Next, we discuss methods of integration in \autoref{sec:integration} and their potential role in explaining \acrshort{lms} in \autoref{sec:explanatory} (\autoref{fig:summary}.3B). Our goal is to provide a focused guide to recent advancements in this field, while addressing current challenges (\autoref{sec:challenges}) and outlining promising directions for future research (\autoref{sec:discussion}).

\section{Related work}
\label{sec:related_work}
In \cite{hollenstein_towards_2020}, the authors review research that utilises various types of cognitive signals acquired during language understanding tasks, focusing on their interplay with \acrfull{ml}-based \acrshort{nlp} methods. Their work discusses both \acrshort{et} and neuroimaging techniques, whereas the present article narrows the focus specifically to \acrshort{et}. Other related reviews include \cite{mathias_survey_2020} and \cite{barrett_sequence_2020}, which explore alternative approaches to the costly and time-intensive data acquisition of gaze behavior at runtime. In contrast, our work examines only the most recent research, reflecting the rapid evolution of the field and the need for an up-to-date review that aligns with broader developments in the field. 
Last, in \cite{cartella_trends_2024}, the authors provide an overview of recent efforts to incorporate human attention mechanisms into 
\acrfull{dl} models. Unlike their review, which offers a broader perspective, this article focuses specifically on language modeling, emphasizing the integration of cognitive signals directly into model architectures rather than solely their applications.



\section{Cognitive signals acquisition}
\label{sec:data}
\subsection{\acrlong{et}} \label{sec:data_et}

When focusing on a stimulus, the eyes perform two key actions: (1) \textit{fixations}, where they pause momentarily to process information, and (2) \textit{saccades}, 
rapid movements during which no visual information is acquired \cite{lopez-cardona_seeing_2024}. The 
interaction of these actions forms what are known as 
\textit{scanpaths}. However, collecting organic \acrshort{et} data presents significant challenges, including the reliance on proprietary, high-precision equipment and data privacy concerns \cite{khurana_synthesizing_2023}. Additionally, the effectiveness of \acrshort{ml} models often hinges on the availability of large datasets for training. Together, these factors often result in insufficient \acrshort{et} data for training \acrshort{ml} models. Compounding this issue, gaze data is typically unavailable during inference.

To address these limitations, several approaches have been proposed. One strategy involves leveraging methods that eliminate the need for gaze data during inference, such as \acrfull{mtl} (see \autoref{subsec:multitask}), or learning modules that develop combined representations (see \autoref{subsec:encoders}). These techniques enable training on \acrshort{nlp} tasks even in scenarios where cognitive data is not availabile for all instances in the primary dataset. Another promising alternative involves the use of generative models designed to predict gaze patterns in response to specific stimuli. This approach not only facilitates the generation of synthetic data for training purposes, but also ensures the availability of gaze-related data during inference (\autoref{fig:integration}.A) As disadvantages, their use may increase the total number of parameters and introduce bias; however, these models are typically much smaller than LMs and can remain frozen during training.

\textbf{Text stimuli.} From raw \acrshort{et} data captured during text reading, various reading measures (often referred to as \acrshort{et} features) can be derived. Among the most widely used features for enhancing language modelling are First Fixation Duration (FFD) and Go-Past Time (GPT). Other metrics, such as Total Reading Time (TRT) and the Number of Fixations (nFix), aggregate the data to provide a comprehensive and robust characterisation of reading behaviour \cite{hollenstein_zuco_2018}. Several \acrshort{et} corpora exist for text, serving both as resources for training models to predict gaze patterns and as datasets for directly integrating organic eye-tracking data into \acrshort{lms}. While English is the dominant language in most available corpora, there is an increasing number of datasets in other languages, especially with the emergence of multilingual initiatives \cite{bolliger_emtec_2024} . Due to space limitations, we do not include a detailed list of these datasets here. Instead, we refer readers to the survey by Mathias et al. \cite{mathias_survey_2020}, which provides a thorough review of datasets available up to 2020. For more recent advancements in \acrshort{et} corpora, the related work section of Bolliger et al. \cite{bolliger_emtec_2024} is an excellent resource.

Models for predicting scanpaths or reading measures from text have undergone significant evolution. Beginning with E-Z Reader \cite{reichle_toward_1998}, one of the first and most influential cognitive models, the field has progressed to more recent approaches that often incorporate \acrshort{lm} such as BERT. For instance, some models predict \acrshort{et} reading metrics at the token level, as demonstrated in \cite{li_torontocl_2021}, and the token-level tasks introduced by \cite{hollenstein_cmcl_2021, hollenstein_cmcl_2022}. Models specifically designed to predict scanpaths, such as Eyettention \cite{deng_eyettention_2023} and ScanTextGAN \cite{khurana_synthesizing_2023}, represent another line of research. These predictive models are often integrated into language modeling workflows, either directly or after fine-tuning. 


\textbf{Images stimuli.} 
In the field of \acrfull{cv}, predictive models of human attention have proven useful for various applications, such as optimising interaction designs and enhancing webpage layouts \cite{li_uniar_2024}. While \acrshort{et} integration in \acrshort{lms} initially focused on scanpaths and reading measures in text, the appearance of linguistic applications involving images, such as \acrshort{vqa} and \acrshort{mllms}, has enabled the incorporation of cognitive data related to images into \acrshort{lms}. Similarly to text-based analysis, specific models have been developed to predict scanpaths in images, with recent examples like TPP-Gaze \cite{damelio_tpp-gaze_2024}.

In addition, saliency prediction models focus on generating heatmaps or probability distributions that highlight areas of attention within an image \cite{cartella_trends_2024}. Recent approaches leverage deep neural networks to automatically learn discriminative features, as demonstrated by models like \cite{chen_learning_2023} and TempSal \cite{aydemir_tempsal_2023}. Despite their utility, a common limitation of these models is their dependence on task-specific configurations, which can restrict their generalisability. Addressing this challenge, UniAR \cite{li_uniar_2024} represents a significant advancement that predicts scanpaths and saliency, but also incorporates user preferences related to the input. By integrating multiple tasks, UniAR improves upon the state-of-the-art in both scanpath and saliency prediction paradigms, outperforming single-task models.
\subsection{Neuroimaging Techniques}
The field of neuroimaging includes a variety of methodologies aimed at measuring brain activity and cognitive processes. Among these, electroencephalography (EEG) stands out for its ability to capture electrical activity from the scalp with high temporal resolution, making it suitable for studying real-time neural dynamics. EEG is a non-invasive and portable method, offering practical advantages over techniques like functional Magnetic Resonance Imaging (fMRI), which excels in spatial resolution, and Magnetoencephalography (MEG), which provides detailed spatiotemporal data but is limited by its high cost and equipment requirements \cite{van2024introduction} \cite{ranasinghe2024functional}.  

In this review, we emphasise the potential of \acrshort{et} data, which offers several advantages over other neuroimaging approaches and its integration into the field of LM has progressed more substantially. \acrshort{et} systems are non-invasive, simple to set up, more affordable, and provide data that is easier to interpret. While the integration of neuroimaging techniques like EEG into \acrshort{lms} lies outside the primary scope of this review, we discuss (\autoref{subsec:app_langunder}) the most relevant works that combine EEG and \acrshort{et}. Additionally, we briefly explore (\autoref{sec:explanatory}) how these techniques contribute to explaining \acrshort{lms} representations.  

\section{\acrshort{lms} augmentation with cognitive signals}
\label{sec:applications}

In the following, we discuss the most relevant studies (\autoref{tab:task_table}) that demonstrate the integration of cognitive signals in \acrshort{lms}, across different \acrshort{nlp} tasks.

\subsection{Language understanding}
    \label{subsec:app_langunder}

        

In language processing, \cite{sood_improving_2020} represents a first effort to integrate reading measures into the attention mechanism of neural networks for tasks like text comprehension, paraphrase generation, and sentence compression. Similarly, \cite{shubi_fine-grained_2024} explored the integration of eye movement features to enhance reading comprehension, employing three distinct paradigms based on the RoBERTa model \cite{liu_roberta_2019}. 
Another area where the integration of human attention has been extensively studied is sentiment classification. Specifically, \cite{yang_plm-as_2023} introduced PLM-AS, a BERT-based model augmented with attention data. Building on this approach, \cite{deng_pre-trained_2023} validated the use of synthetic scanpaths by applying the same methodology to the General Language Understanding Evaluation (GLUE) benchmark, which spans nine language understanding tasks. More recently, \cite{deng_fine-tuning_2024} extended this work by training two parallel modules during fine-tuning---one augnented with scanpaths and the other without---to develop language representations grounded in human cognitive processing. Similarly, \cite{wang_gaze-infused_2024} and \cite{ding_cogbert_2022} demonstrated the benefits of incorporating reading measures into BERT’s training, modifying the Transformer attention mechanism to improve performance on GLUE benchmark tasks.  

In a related effort, \cite{ren_cogalign_2021} introduced CogAlign, a multi-task neural network designed to align neural text representations with cognitive processing signals across various NLP tasks. Using an adversarially-trained shared encoder, CogAlign facilitates the transfer of cognitive knowledge to datasets lacking recorded cognitive signals, incorporating both \acrshort{et} and EEG data. Expanding on this work, \cite{zhang_cogaware_2024} developed CogAware, an advanced architecture tailored to learn cross-domain features while simultaneously preserving modality-specific characteristics. These studies, alongside earlier research integrating EEG and \acrshort{et} data, highlight the successful integration of \acrshort{et} with other cognitive signals using the popular ZUCO dataset \cite{hollenstein_zuco_2018}, a comprehensive resource combining EEG and \acrshort{et} signals.

\begin{table}[t!]
    \caption{Summary of Reviewed Works classify by Applications (\autoref{sec:applications}), Nature of data: Text (T), Images (I)) and source (\autoref{sec:data_et}), and the integration techniques employed (\autoref{sec:integration}).}
    \begin{center}
    \begin{adjustbox}{max width=\textwidth}
    \scriptsize
    \begin{tabular}{cccccc}
    
        \hline
        \textbf{} & \textbf{Application} & \textbf{T} & \textbf{I} & \textbf{Data} & \textbf{Integration} \\
        
        \hline
        \cite{sood_improving_2020} & Text Comprehension & \checkmark &   & \cite{reichle_toward_1998} & Attention \\
        \cite{shubi_fine-grained_2024} & Reading Comprehension & \checkmark &   & Real & Input, Cross Att. \\
        \cite{yang_plm-as_2023} & Sentiment Classification & \checkmark &   & Real & Reorder \\
        \cite{deng_pre-trained_2023} & Sentiment Classification & \checkmark &   & \cite{deng_eyettention_2023} & Reorder \\
        \cite{deng_fine-tuning_2024} & Language Understanding & \checkmark &   & \cite{deng_eyettention_2023} & Reorder \\
        \cite{wang_gaze-infused_2024} & Language Understanding & \checkmark &   & \cite{li_torontocl_2021} & Transformer Att. \\
        \cite{ding_cogbert_2022} & Language Understanding & \checkmark &   & Self-built & Transformer Att. \\
        \cite{ren_cogalign_2021} & Language Understanding & \checkmark &   & Real & Encoder \\
        \cite{zhang_cogaware_2024} & Language Understanding & \checkmark &   & Real & Encoder \\
        \cite{yu_ceer_2024} & Name Entity Recognition & \checkmark &   & Real & Encoder \\
        \hline
        \cite{huang_longer_2023} & Language Modeling & \checkmark &   & Self-built & Architecture \\
        \cite{huang_long-range_2023} & Language Modeling & \checkmark &   & Self-built & Architecture \\
        \cite{lopez-cardona_seeing_2024} & Human Alignment & \checkmark &   & \cite{li_torontocl_2021} & Input \\
        \cite{kiegeland_pupil_2024} & Human Alignment & \checkmark &   & \cite{deng_eyettention_2023} & Reorder \\
        \hline
        \cite{malmaud_bridging_2020} & Question Answering & \checkmark &   & Real & \acrshort{mtl} \\
        \cite{zhang_eye-tracking_2024} & Question Answering & \checkmark &   & \cite{li_torontocl_2021} & Attention Mask \\
        \cite{sood_multimodal_2023} & Visual Question Answering & \checkmark & \checkmark & \cite{sood_improving_2020} & Transformer Att. \\
        \cite{inadumi_gaze-grounded_2024} & Visual Question Answering &   & \checkmark & Real & Input \\
        \cite{yan_voila-_2024} & Visual Question Answering &   & \checkmark & \cite{openai_gpt-4_2023} & Attention \\
        \hline
        \cite{maharaj_eyes_2023} & Hallucination Detection & \checkmark &   & Self-built & Attention \\
        \cite{alacam_eyes_2024} & Hate Speech Detection & \checkmark &   & Real & Input \\
        \cite{tiwari_predict_2023} & Sarcasm Detection & \checkmark &   & Self-built & Cross Att. \\
        \cite{khurana_synthesizing_2023} & Sarcasm Detection & \checkmark &   & Self-built & Input \\
        \bottomrule
    \normalsize
    \end{tabular}
    \end{adjustbox}
    \end{center}
    \label{tab:task_table}
\end{table}

\subsection{Language modeling}
    \label{subsec:app_lm}

Language modelling is a fundamental \acrshort{nlp} task centered on predicting the next word or token in a sequence. In \cite{huang_longer_2023}, TRT was incorporated into a parallel \acrfull{rnn} structure for \acrshort{lm}. Building on this, \cite{huang_long-range_2023} employed TRT to manage the cache of a Transformer-based architecture, demonstrating its effectiveness with BERT.
    
\subsection{Language Models alignment}
    \label{subsec:app_alig}


\acrshort{llms} are typically trained on massive datasets and often require extensive fine-tuning to align their outputs with human expectations. A significant body of research has focused on refining how \acrshort{llms} interpret and respond to user intent \cite{wang_aligning_2023}, leading to the development of human alignment techniques. A widely adopted strategy for achieving human alignment involves leveraging explicit human feedback as preference data. Established methodologies in this area include \acrfull{rlhf} \cite{ouyang_training_2022} and \acrfull{dpo} \cite{rafailov_direct_2023}. The first study to integrate explicit feedback with implicit gaze-based feedback is \cite{kiegeland_pupil_2024}, which proposed a novel dataset generation method for \acrshort{dpo}. This work builds on earlier sentiment generation frameworks \cite{deng_pre-trained_2023, yang_plm-as_2023} that incorporated \acrshort{et} for sentiment classification. Additionally, \cite{lopez-cardona_seeing_2024} integrated \acrshort{et} reading measures within the \acrfull{rm} of the \acrshort{rlhf} framework \cite{dubey_llama_2024}.

\subsection{\acrfull{qa}}
    \label{subsec:app_qa}
    
        
        

Malmaud et al. \cite{malmaud_bridging_2020} enhanced a standard Transformer architecture for multiple-choice \acrshort{qa} by simultaneously predicting human reading times for text stimuli. Following this, \cite{zhang_eye-tracking_2024} improved performance on the SQuAD benchmark by masking Transformer attention using gaze data. 
When \acrshort{qa} involves questions that require visual information, it is referred to as \acrshort{vqa}, an important task that bridges \acrshort{cv} and \acrshort{nlp}. 
Sood et al. \cite{sood_multimodal_2023} introduced the first method for multimodal integration of human-like attention across images and text. More recently, Inadumi et al. \cite{inadumi_gaze-grounded_2024} utilised gaze-highlighted image regions to enhance \acrshort{vqa} capabilities. Similarly, Yan et al. \cite{yan_voila-_2024} leveraged trace data as a proxy for gaze information, incorporating it into training paradigms for \acrshort{mllms}, highlighting the growing importance of gaze and visual data in \acrshort{qa} systems.  

\subsection{Other \acrshort{nlp} tasks}
    \label{subsec:app_hallu}



Human attention has also been applied to improve other \acrshort{nlp} tasks. For instance, in hallucination detection, \cite{maharaj_eyes_2023} used element-wise multiplication of token reading times with BERT token embeddings before the final prediction layer. Moreover, \cite{alacam_eyes_2024} demonstrated improvements in hate speech detection by concatenating \acrshort{et} reading measures with token embeddings. 
Gaze behaviour has also proven effective for sarcasm detection, a task closely related to sentiment analysis (\autoref{subsec:app_langunder}). For example, \cite{tiwari_predict_2023} combined \acrshort{et} features in a multimodal architecture with video and text inputs. Similarly, \cite{khurana_synthesizing_2023} integrated generated scanpaths with BERT embeddings for binary sarcasm detection. 

\section{Cognitive signals integration into \acrshort{lm}} 
\label{sec:integration}


\begin{figure*}[h]
    \begin{center}
    \includegraphics[width=0.98\linewidth]{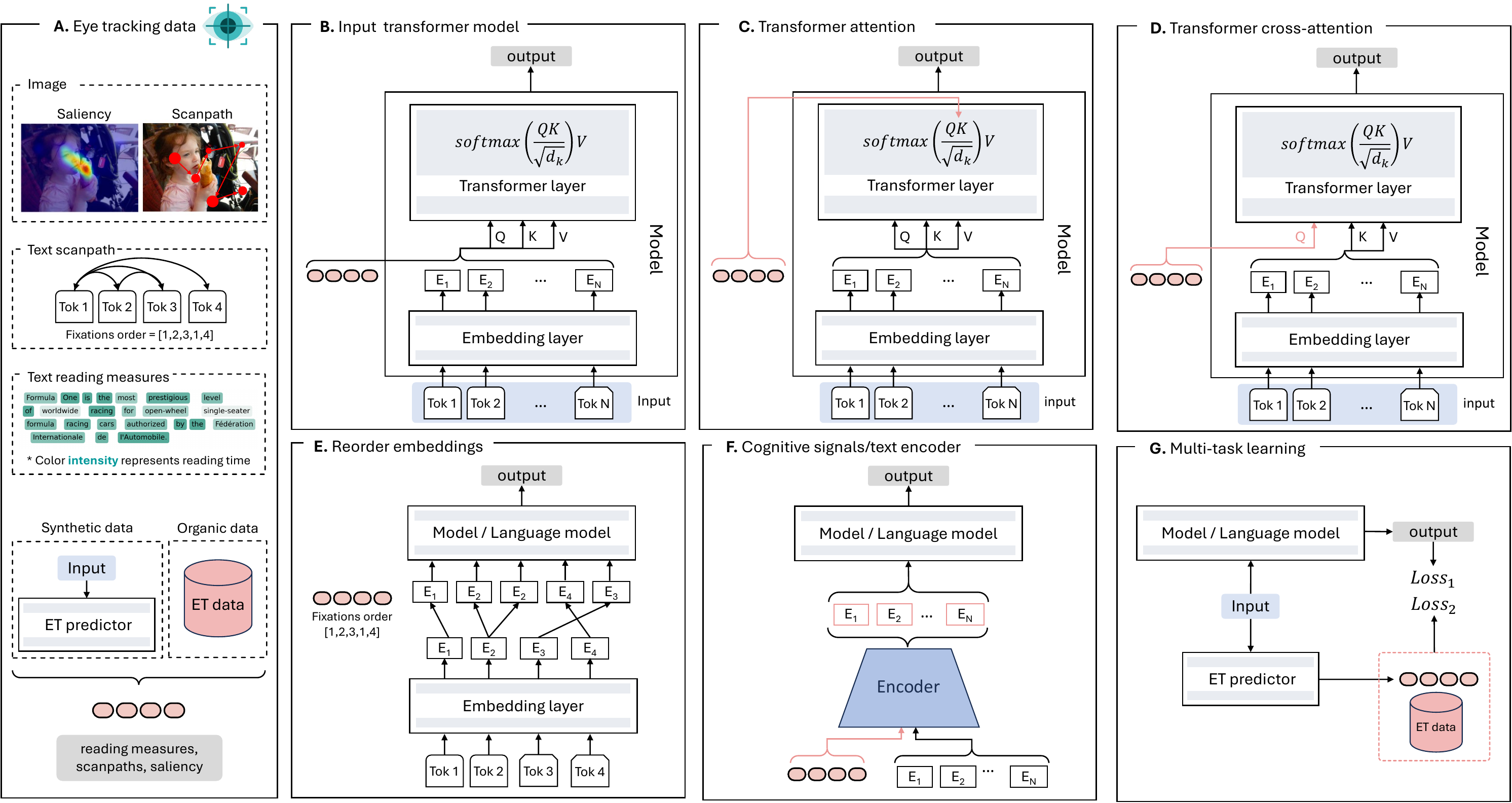}
    \end{center}
    \caption{Schemas of different processes of integrating ET information into \acrshort{lms} covered in \autoref{sec:integration}.}
    \label{fig:integration}
\end{figure*}
In what follows, we categorise the studies outlined in \autoref{tab:task_table} and \autoref{sec:applications} according to criteria such as the type of data utilised, how this data is integrated into \acrshort{lm}, the source of the data, and, additionally, by considering how 
alignment is achieved. Specifically, some methods leverage scanpaths, as seen in studies like \cite{yang_plm-as_2023, deng_pre-trained_2023, deng_fine-tuning_2024, kiegeland_pupil_2024, khurana_synthesizing_2023}. However, the majority of models analysed in this review primarily rely on reading measures that assess data at the level of words or tokens. The three approaches that incorporates images \cite{sood_multimodal_2023}, \cite{inadumi_gaze-grounded_2024}, and \cite{yan_voila-_2024} leverage saliency maps.

Approaches incorporating cognitive signals during inference can be categorised into: (1) those utilising organic data \cite{malmaud_bridging_2020, ren_cogalign_2021, zhang_cogaware_2024, alacam_eyes_2024, shubi_fine-grained_2024, yang_plm-as_2023, yu_ceer_2024, inadumi_gaze-grounded_2024}, and (2) those relying on generative models to predict cognitive signals. The latter group includes methods that develop their own predictive models 
\cite{khurana_synthesizing_2023, ding_cogbert_2022, maharaj_eyes_2023, huang_long-range_2023, huang_longer_2023, tiwari_predict_2023}. Alternatively, some studies employ pre-trained models
\cite{zhang_eye-tracking_2024, deng_fine-tuning_2024, wang_gaze-infused_2024, sood_multimodal_2023, deng_pre-trained_2023, lopez-cardona_seeing_2024, kiegeland_pupil_2024}. For example, \cite{sood_improving_2020} developed predictive models using data generated by E-Z Reader \cite{reichle_toward_1998}. Also, \cite{yan_voila-_2024} introduced an innovative use of \acrshort{llms}, specifically GPT-4 \cite{openai_gpt-4_2023}, to annotate labels for predicting \acrshort{et} heatmaps in images for the first time.

The integration of cognitive signals into \acrshort{lms} can be achieved through various methods, addressed next. \autoref{fig:integration} provides a schematic overview of several integration options. While this representation is designed to be general, it is important to acknowledge that the specifics of these integrations may vary depending on the approach or the study in question.

\subsection{Model input}
     \label{subsec:model_input}
A popular approach involves modifying text embeddings by integrating gaze information. For example, RoBERTa-QEye \cite{shubi_fine-grained_2024} concatenates \acrshort{et} data as additional input features. Other works, such as \cite{alacam_eyes_2024, lopez-cardona_seeing_2024, khurana_synthesizing_2023}, follow a similar approach, while GazeReward \cite{lopez-cardona_seeing_2024} modifies embeddings directly. A representative example of this approach is shown in \autoref{fig:integration}.B. For \acrshort{vqa}, \cite{inadumi_gaze-grounded_2024} extracted areas of interest from images using gaze-source heatmaps and combineed this information with the image as input to the model.

Some approaches focus on reordering contextual word or token embeddings (\autoref{fig:integration}.E) based on fixation sequences observed in the text \cite{yang_plm-as_2023, deng_pre-trained_2023, kiegeland_pupil_2024}. For instance, MAGQEye \cite{shubi_fine-grained_2024} incorporates \acrshort{et} data to modify contextualised word representations. Similarly, \cite{deng_fine-tuning_2024} reordered embeddings and created 
a BERT-based architecture using these reordered embeddings, training their model by minimising the sum of both losses.

\subsection{Modifing model representation}
    \label{subsec:model_representation}
The introduction of attention layers further enabled the integration of \acrshort{et} data, by leveraging it as a proxy for human attention. 
In particular, \cite{sood_improving_2020} incorporated gaze information into their Task Model alongside the original input sentence to generate an output sentence. Another example is \cite{maharaj_eyes_2023}, which used \acrshort{et} data as attention signals, applying element-wise multiplication with BERT token embeddings before the final prediction layer. 
In the context of \acrshort{vqa}, \cite{yan_voila-_2024} extended this concept to images by creating gaze patches, which are further refined into key embeddings. These embeddings are then modulated through a gating mechanism designed to incrementally incorporate gaze data. The resulting gaze and image key embeddings are combined and processed through a self-attention mechanism, synthesising the information into a unified set of latent perceiver embeddings.

\textbf{Transformer attention.} 
As Transformer-based language models continue to advance, researchers have increasingly explored the integration of gaze information into the self-attention mechanism of the Transformer (\autoref{fig:integration}.C). This mechanism is designed to efficiently capture global dependencies by computing relationships within a sequence, relating each element to all others. It achieves this by generating Query ($Q$), Key ($K$), and Value ($V$) vectors for each element, calculating relevance scores through dot products, and producing updated representations (\autoref{eq:attention}).

 \begin{equation}
    \text{Attention}(Q, K, V) = \text{softmax}\left(\frac{QK^\top}{\sqrt{d_{k}}}\right) V
    \label{eq:attention}
\end{equation}

In \cite{wang_gaze-infused_2024}, the authors extend \autoref{eq:attention} by incorporating gaze representations into the key vectors, thereby enriching the attention mechanism with cognitive signals. Similarly, \cite{ding_cogbert_2022} built on this idea by introducing gaze-guided enhanced attention, which assigns varying weights to cognitive features across different layers of the model. In the context of multimodal language models for \acrshort{vqa}, \cite{sood_multimodal_2023} proposed an innovative attention function applied in select layers. This function multiplies human-like attention weights, denoted as $\alpha$, with the attention scores derived from \autoref{eq:attention}. The resulting Transformer-based architecture is designed to process both images and text, and it is the first study to integrate gaze-based attention across these modalities.

\textbf{Cross attention.} Cross-attention is a powerful mechanism that computes interactions between two distinct sequences, typically referred to as $Q$ and $K,V$ pairs (\autoref{fig:integration}.D). Unlike self-attention, where $Q$, $K$, and $V$ are derived from the same sequence, cross-attention enables the model to relate information across different modalities or inputs
(\autoref{eq:attention}). For example, \cite{tiwari_predict_2023} integrated gaze-based features alongside other modalities (e.g., text, audio, video). A collaborative gating mechanism was employed to fuse these diverse features, enabling label prediction through a feed-forward neural network. Similarly, the PostFusion-QEye model \cite{shubi_fine-grained_2024} processes text and eye movement data separately, merging them later through cross-attention mechanisms to enhance performance.

\textbf{Masking.} Masking is a common technique in Transformer-based architectures, ensuring that each token attends only to specific tokens. This is typically implemented using binary values 
to indicate whether a token should be included in the attention computation. In \cite{zhang_eye-tracking_2024}, the authors proposed an innovative approach to determine these masking values dynamically, basing them on \acrshort{et} features for each token. This method enriches the \acrshort{lm}'s performance by incorporating cognitive signals into the masking process.

\subsection{Encoders}
    \label{subsec:encoders}
Alternatively, some models are designed to encode both textual and cognitive signals
(\autoref{fig:integration}.F). During inference, these models require only text input, allowing the learned relationships to be applied without the need for cognitive input, as demonstrated in \cite{ren_cogalign_2021} and \cite{zhang_cogaware_2024}. Another example is \cite{yu_ceer_2024}, where the authors trained an encoder to represent words as high-dimensional embeddings and decode them into gaze feature vectors. This approach captures gaze-related features that reflect the 
attention words receive,
thereby improving the encoder's ability to model human-like reading processes.

\subsection{\acrlong{mtl}}
    \label{subsec:multitask}
\acrshort{mtl} is an established \acrshort{ml} paradigm that improves the performance of a primary task by simultaneously optimising auxiliary tasks. Research has indicated that learning gaze behaviour can significantly improve the performance of \acrshort{nlp} tasks, even when gaze data is not required during inference (\autoref{fig:integration}.G). For example, \cite{malmaud_bridging_2020} presented a multi-task reading comprehension architecture where a state-of-the-art Transformer model jointly addressed \acrshort{qa} and predicted the distribution of human reading times across the text.
    
\subsection{Modifying model architecture}

Another line of research leverages \acrshort{et} metrics to inform architectural design in Transformer-based models. For instance, \cite{huang_long-range_2023} used the TRT of each token to select essential key-value pairs for storage in the Transformer's cache, optimizing memory usage. Similarly, \cite{huang_longer_2023} introduced a parallel \acrshort{rnn} structure where fixation duration determined the workload by controlling the number of activated parallel \acrshort{rnn} units.
\section{Explaining \acrshort{lms} with cognitive signals}
\label{sec:explanatory}

The significant applications of \acrshort{llms} 
have motivated research into their internal representations and their resemblance to human language processing in the brain \cite{zhang_word_2024}. Recent findings in neuroscience and \acrshort{ai} reveal a commonality: both biological brains and artificial models generate similar representations when exposed to comparable stimuli. In \autoref{subsec:exp_attention}, we discuss research on human attention and its parallels with attention mechanisms of \acrshort{lms} in language processing. Also, \autoref{subsec:exp_neuroimages} provides a concise overview of recent advancements in this field using neuroimaging techniques with a summary of all reviewd works in \autoref{tab:explanatory}.

\subsection{Human- vs. model-based relative attention comparison} \label{subsec:exp_attention}

The relationship between cognitive signals during reading and attention mechanisms in \acrshort{lms} has received a lot of attention (\autoref{fig:summary}.3B), with many studies attempting to enhance cognitive interpretability and explainability in \acrshort{lms} \cite{deng_eyettention_2023, bolliger_emtec_2024}. Several architectures have been proposed, with a focus on Transformer-based frameworks. Examples include studies analysing encoder models like BERT and RoBERTa \cite{hollenstein_relative_2021, wang_gaze-infused_2024, bensemann_eye_2022}, decoder-based models like GPT \cite{wu_eye_2024, wang_probing_2024}, and encoder-decoder architectures such as T5 \cite{eberle_transformer_2022}. 
These works employ gaze reading metrics, such as TRT \cite{hollenstein_relative_2021}, \cite{eberle_transformer_2022}, alongside other measures \cite{sood_interpreting_2020, wang_gaze-infused_2024, wang_probing_2024}. Specifically, prior studies have examined attention mechanisms across different model layers, with some focusing on early layers \cite{hollenstein_relative_2021, bensemann_eye_2022}, others on final layers \cite{sood_interpreting_2020}, and several on multi-layer analysis \cite{wang_gaze-infused_2024}. Comparative techniques such as Spearman correlation \cite{wang_gaze-infused_2024, bensemann_eye_2022} and KL divergence \cite{sood_interpreting_2020} have revealed diverse alignment patterns between human and model attention. 
Alternative attention metrics have also been explored. For instance, flow attention has been examined in \cite{eberle_transformer_2022}, while gradient-based saliency methods have been studied in studies such as \cite{hollenstein_relative_2021, wu_eye_2024}, yielding promising results. Furthermore, research into the alignment of a model's training objectives with tasks used while collection \acrshort{et} corpora has revealed complex relationships in attention patterns \cite{wu_eye_2024, sood_interpreting_2020, eberle_transformer_2022}.

\subsection{LMs interpretion through Neuroimaging Techniques}
\label{subsec:exp_neuroimages}


Recently, researchers have used advanced \acrshort{lm} to explore neural activity during language processing. Numerous studies have highlighted structural similarities between brain activations and those of language models, enabling linear partial mappings between features from neural recordings and computational models (\autoref{fig:summary}.3B). In \cite{karamolegkou_mapping_2023}, the authors review over 30 studies published prior to 2023. Their analysis suggests that, although the evidence is still unclear, correlations with model size and quality provide cautious optimism.
Later, \cite{han_investigating_2024} examined how modality and training objectives influence representational alignment between Transformers and brain activity. Additionally, \cite{antonello_evidence_2024} established a connection between fMRI data and abstraction processes in \acrshort{lms}, while \cite{nakagi_unveiling_2024} presented evidence of multi-level and multi-modal semantic representations in human brain using \acrshort{llms}.

\begin{table}[t!]
    \caption{Summary of Reviewed Works classify by cognitive signal (\autoref{sec:data_et}), and the features used of each model.}
    \begin{center}
    \begin{adjustbox}{max width=\textwidth}
    \scriptsize
    \begin{tabular}{cccc}
    
        \hline
         & \textbf{Cognitive signal} & \textbf{Model} & \textbf{Model features} \\
   \hline
    
    \cite{eberle_transformer_2022} & ET & BERT based, T5, other & att. final layer, att. flow \\
    \cite{bensemann_eye_2022} & ET & BERT based, other & att. first layer \\
    \cite{sood_interpreting_2020} & ET & other & att. last layer \\
    \cite{wang_gaze-infused_2024} & ET & BERT based & att. all layers \\
    \cite{wang_probing_2024} & ET & GPT based & att. all layers, output \\
    \cite{wu_eye_2024} & ET & BERT based, GPT based & saliency per token \\
    \cite{hollenstein_relative_2021} & ET & BERT based & saliency per token \\
    \hline
    \cite{han_investigating_2024}& fMRI & Llama, LlaVA, ResNet & hidden states\\
    \cite{antonello_evidence_2024}& fMRI & OPT & hidden states\\
    \cite{nakagi_unveiling_2024}& fMRI & Llama, OPT, BERT, GPT& hidden states\\
    \bottomrule
    \normalsize
    \end{tabular}
    \end{adjustbox}
    \end{center}
    \label{tab:explanatory}
\end{table}
\section{Open challenges and future directions}
\label{sec:challenges}
As discussed in \autoref{sec:data_et}, an emerging challenge is \acrshort{et} data scarcity, which translates to two areas for improvement: (1) generating more and richer data representations, and (2) developing more sophisticated models that can predict human behaviour. Regarding the former, newly created datasets like WebQAmGaze \cite{ribeiro_webqamgaze_2023}, which collect gaze data via webcams, offer promising solutions for simplifying data acquisition. Advances in these technologies may realise a paradigm shift in how data is acquired and used for inference. An alternative approach for visual stimuli involves using mouse clicks \cite{kim_bubbleview_2017} or mouse cursor trajectories \cite{arapakis2020} that can be used as proxies of visual attention. \cite{yan_voila-_2024} is another example of work that proposes the generation of eye-tracking datasets using said method, annotated with GPT4. Regarding the latter, there is significant potential for enhancing models that predict gaze behavior in response to stimuli, whether textual or visual. Although progress has been made in model architectures, the primary limitation remains the lack of sufficient (or sufficiently rich) training data. Addressing this bottleneck could unlock advancements, as demonstrated by the UniAr \cite{li_uniar_2024} model.




\section{Discussion}
\label{sec:discussion}

Traditionally, \acrshort{et} applications have been used for language comprehension, including tasks such as sentiment classification, sarcasm detection, and reading comprehension. Today, the challenges faced by \acrshort{llms} and \acrshort{mllms} include alignment with human intentions, hallucinations, seamless multimodal integration \cite{casper_open_2023}, data bottleneck and training costs \cite{villalobos_will_2024, erdil_data_2024}. Among these, alignment---ensuring that models act ethically and produce outputs consistent with human values---remains a critical issue, as biases and harmful outputs persist. Similarly, hallucinations, particularly in \acrshort{mllms}, erode trust by generating false or misleading information \cite{yin_survey_2024}.

As outlined in \autoref{sec:applications}, recent research highlights the potential of integrating cognitive signals into new applications to tackle these emerging challenges. For instance, enriching existing datasets with user-derived cognitive signals could help alleviate the data bottleneck problem by addressing data scarcity. When it comes to human alignment---capturing user preferences through explicit feedback---cognitive signals provide valuable insights, as discussed in \autoref{subsec:app_alig}. For example, eye movement patterns reveal the information humans prioritise during decision making \cite{adams_active_2015}, which in turn affects attention allocation in preference judgments. Models like UniAR \cite{li_uniar_2024} further demonstrate the connection between visual attention and explicit feedback, such as aesthetic evaluations or subjective preferences. This makes the integration of cognitive signals particularly promising for advancing this challenge.

Moreover, the integration of vision and language in \acrshort{mllms} has enabled additional opportunities for leveraging cognitive signals, extending beyond text-based methods to include images and bridging diverse forms of human communication. This innovation has driven advancements in applications like \acrshort{vqa} \cite{sood_multimodal_2023}. Within this framework, \acrshort{et} in particular appears to be a promising technique for enhancing model performance, specifically in addressing hallucinations. Research indicates that incorporating  \acrshort{et} can mitigate such issues in text-based models \cite{maharaj_eyes_2023}, and their recent integration into multimodal frameworks, as shown by \cite{yan_voila-_2024}, paves the way for further advancements.

Drawing on the parallels between the representations of \acrshort{lms} and the human brain discussed in \autoref{sec:explanatory}, we argue that the analysis in this review highlights the significant potential of integrating cognitive data into \acrshort{lms}. With the rapid advancements in this field, such integration could play a pivotal role in addressing key challenges like human alignment and hallucination detection in \acrshort{mllms}. Looking ahead, we expect these models to benefit greatly from the growing trend of incorporating data sources such as \acrshort{et}.
\textbf{Ethical statement.} It is crucial to acknowledge and address potential privacy issues and biases in cognitive data collection and processing by ensuring unlinkability to individual participants. \cite{kiegeland_pupil_2024} \cite{deng_fine-tuning_2024}. Furthermore, it is equally important to consider and mitigate potential biases \cite{kiegeland_pupil_2024}.


\bibliographystyle{IEEEtran}

\bibliography{bib/references_manual}

\end{document}